\documentclass[letterpaper, 10 pt, conference]{ieeeconf}
\IEEEoverridecommandlockouts
\usepackage{cite}
\usepackage{amsmath,amssymb,amsfonts}
\usepackage{algorithmic}
\usepackage{graphicx}
\usepackage{textcomp}
\usepackage{xcolor}
\usepackage{url}
\usepackage{hyperref}

\def\BibTeX{{\rm B\kern-.05em{\sc i\kern-.025em b}\kern-.08em
    T\kern-.1667em\lower.7ex\hbox{E}\kern-.125emX}}
\begin{document}

\title{\LARGE\bfseries
Lasso Gripper: A String Shooting–Retracting Mechanism\\[-0.4ex]
for Shape-Adaptive Grasping
}

\author{Qiyuan Qiao$^*$, Yu Wang$^*$, Xiyu Fan, and Peng Lu%
\thanks{The hardware design specifications and associated control algorithms required to replicate the experimental results have been deposited in an open-source repository: \url{https://github.com/qiaoqy/lasso_gripper}}
\thanks{$^*$ These two authors contributed equally to this research.}
\thanks{This work was supported by General Research Fund under grant no. 17204222, and in part by the Seed Fund for Collaborative Research and General Funding Scheme-HKU-TCL Joint Research Center for Artificial Intelligence.}
\thanks{The authors are with the Adaptive Robotic Controls Lab (ArcLab),
Department of Mechanical Engineering, The University of Hong Kong,
Hong Kong SAR, China.
(email: qiaoqy, ywang812, fanxiyu@connect.hku.hk; lupeng@hku.hk) }
}

\maketitle

\begin{abstract}
Handling oversized, variable-shaped, or delicate objects in transportation, grasping tasks is extremely challenging, mainly due to the limitations of the gripper's shape and size. This paper proposes a novel gripper, Lasso Gripper. Inspired by traditional tools like the lasso and the uurga, Lasso Gripper captures objects by launching and retracting a string. Contrary to antipodal grippers, which concentrate force on a limited area, Lasso Gripper applies uniform pressure along the length of the string for a more gentle grasp. The gripper is controlled by four motors—two for launching the string inward and two for launching it outward. By adjusting motor speeds, the size of the string loop can be tuned to accommodate objects of varying sizes, eliminating the limitations imposed by the maximum gripper separation distance. To address the issue of string tangling during rapid retraction, a specialized mechanism was incorporated. Additionally, a dynamic model was developed to estimate the string's curve, providing a foundation for the kinematic analysis of the workspace. In grasping experiments, Lasso Gripper, mounted on a robotic arm, successfully captured and transported a range of objects, including bull and horse figures as well as delicate vegetables. The demonstration
 video is available here: \url{https://youtu.be/PV1J76mNP9Y}
\end{abstract}
 
\section{INTRODUCTION}

In the field of robotics, the capability to swiftly and securely grasp and manipulate objects is crucial, particularly in uncharted natural environments or in highly demanding industrial production settings. The limitations of the conventional types of gripper, i.e. the antipodal point grippers, often occurred in handling variable objects with unknown properties. For example, to handle a delicate target, additional sensors or degrees of freedom are needed to prevent the target from slippery between the two fingers and being crushed with excessive forces, reducing their efficiency in dynamic and challenging applications\cite{kim_development_2007, izumi_design_2024, tisdale2023fractal, yin_closing_2020, do_densetact-mini_2023}. Another significant issue in grasping is that the size of the stationed robotic arm profoundly influences the effective reach \cite{10081306}. The demand for more versatile and effective gripping solutions has become increasingly urgent.

\begin{figure}[t]
    \centering
    \includegraphics[width=0.9 \linewidth]{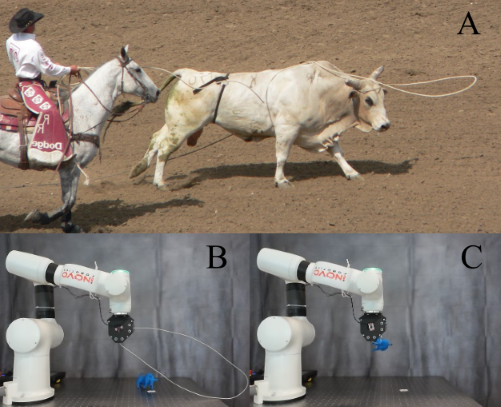} 
    \caption{Traditional lasso and the proposed \textbf{Lasso Gripper}. (A) Traditional lasso is used to catch cattle. Benefiting from the use of horn structure, the loop structure can afford the large pulling force \cite{california_lasso}. (B) The string shot by the Lasso Gripper maintains a stable self-supporting loop structure in the air. (C) As the Lasso Gripper is triggered, the string loop structure caught the horns of the bull model.}
    \label{overall}
\end{figure}

Several investigations into soft actuators have been conducted to address these challenges. Notably, the enveloping grasping technique distinguishes itself through its enhanced adaptability and stability. Various underactuated manipulators with different configurations, capable of modulating both stiffness and shape, have also been explored. \cite{hu2023dual, chen2024soft, zhang2017design, yong_design_2024, sun2024low, ciocarlie2014velo}. Moreover, through advancements in material and micro-structural technology, significant progress has been made in altering the physical properties of flexible materials like ropes. By modifying the mechanical properties of the material either internally or externally, novel physical performance characteristics can be utilized to grasp objects\cite{wang2021lightweight}\cite{olukayode_design_2023}. Beyond mechanical inspirations, many researchers are looking to nature for solutions to object grasping. Drawing inspiration from animals and insects, innovative gripper designs have demonstrated superior performance compared to traditional parallel grippers\cite{sun2020soft}\cite{li2017development}. However, few studies have proposed an omnipotence design to handle objects both exceed the maximum separation distance of the gripper and sensitive to applied force.

To enhance the ability to handle oversized, variable-shaped or delicate objects, Lasso Gripper is proposed as a solution, as illustrated in Fig. \ref{overall}. Inspired by traditional tools like the lasso and the uurga, Lasso Gripper is designed to overcome the limitations of existing grippers by combining a wide operational workspace with the capacity of substantial loads and the adaptability to different object shapes.

A key feature of Lasso Gripper is its ability to achieve a large but adjustable initial capture range, which significantly increases the likelihood of successful grasping. The string loop can ensnare objects within this larger capture range, making Lasso Gripper more tolerant of uncertainties in the object's position. This adaptability allows the gripper to accommodate errors in the position estimation at the moment of grasping, thereby increasing its effectiveness in dynamic environments. Moreover, by adjusting the capture range, Lasso Gripper can capture objects in motion, including those that are airborne. This flexibility makes it particularly suitable for a wide range of applications.

Furthermore, Lasso Gripper is well-suited for handling heavier loads because the string loop applies a direct tensile force to secure the object. In contrast, parallel-finger grippers rely on friction to maintain their grip, which may be less effective for heavier items. This fundamental distinction in the gripping mechanism enables Lasso Gripper to hold heavier objects with greater reliability.

The main contributions of this research are as follows: we proposed a novel string loop grasping mechanism, established the dynamics and kinematic models for the lasso, and conducted the verification demonstrations under different scenarios.

The rest of the paper is organized in the following order: Section II covers the structure and mechanism of Lasso Gripper. Section III discusses the dynamics and kinematics of the lasso both in the air and upon contact with the target. Section IV details the experiment. Section V concludes the study with a discussion of the experiment and further studies.

\section{DESIGN AND PRINCIPLE}

Taking inspiration from traditional tools and popular toys, the string loop shooting structure is modified for grasping tasks. The mechanical structure of the string launching and retraction is designed. The circuit control system is also adjusted to achieve the grasping effect.

\begin{figure}[htbp]
    \centering
    \includegraphics[width=0.8 \linewidth]{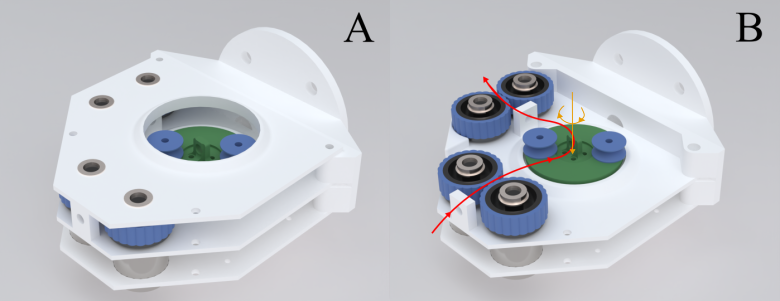} 
    \caption{Overview of the System: (A) Lasso Gripper with top cover; (B) Detail View of the system. The larger blue wheels are the launching mechanism. The green wheel indicates the retracting mechanism, whose rotation direction in the retracting procedure is noted as the yellow rotation vector.}
    \label{mech_illu}
\end{figure}

\subsection{Structure design of Lasso Gripper}

The design of Lasso Gripper draws inspiration from traditional tools like the lasso and the uurga, both of which are manually operated by rotating the tool to launch the noose toward the target and retracting it by pulling the cable. With the popularity of some YouTube videos, the string shooter configuration has been noted to have great potential in the field of grasping \cite{youtube1, youtube2, fazio2024string}. To replicate this functionality in an automated system, our lasso gripper is equipped with four coreless motors and one brushless motor. The gripper’s structure includes a 3D-printed baseboard, where the motors are mounted. The core idea of the baseboard is to be both lightweight and durable to maintain structural integrity while keeping the overall system light. The motors are mounted on slotted brackets, allowing their positions to be adjusted to accommodate different cable diameters. The shape of the loop can be adjusted by increasing the speed of the launch wheels to change the shooting speed of the string.

\begin{figure}[htbp]
    \centering
    \includegraphics[width=0.8 \linewidth]{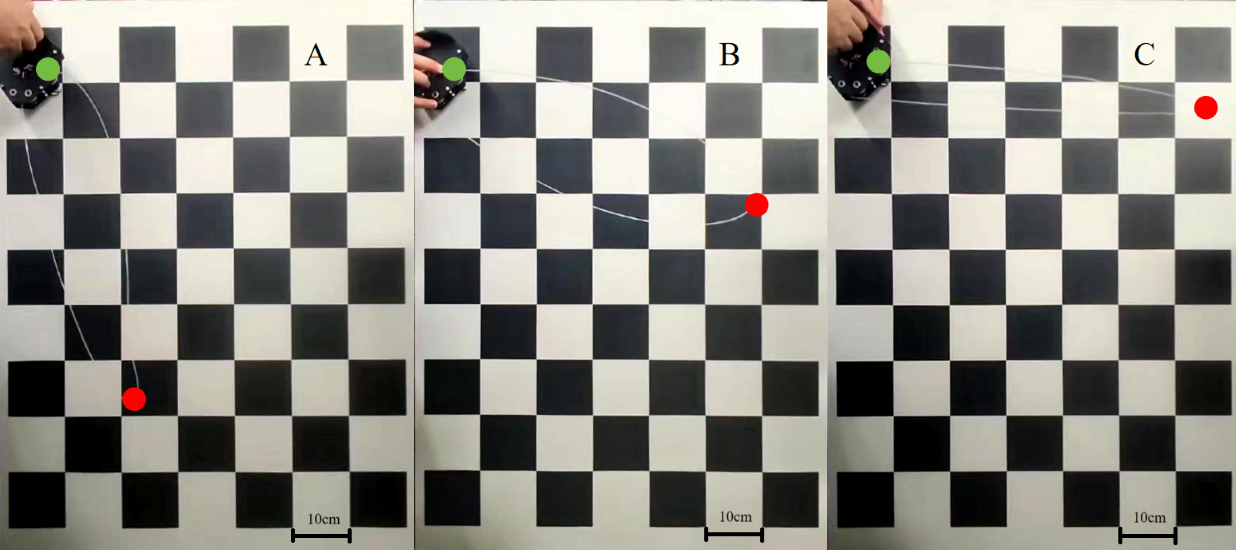} 
    \caption{Increasing the speed of the driving wheels from (A) to (C), the shape of the string loop can be adjusted. The ejection point and the furthest point are noted as green and red points on the string loop. The length of the square side on the chessboard is 10cm.}
    \label{speed}
\end{figure}

To ensure sufficient friction between the string and the launch wheels, customized Thermoplastic Polyurethane (TPU) wheels are used. These blue TPU wheels, visible in the design in Fig. \ref{mech_illu}, are critical for maintaining control over the string during both launch and retraction. The elasticity and durability of TPU make it an ideal material for these wheels, providing consistent grip and reducing wear over time. To minimize unwanted friction between the string and other components of the shooter, ball bearings are incorporated into the winding mechanism and the wheels. These bearings help ensure smooth operation, reducing the mechanical resistance that could otherwise hinder performance.

\subsection{Mechanism}

The control system of Lasso Gripper is managed by an ESP32, which serves as the main Micro-Controller Unit (MCU) for the five motors. This MCU also receives command signals from the upper computer to manage the launch, maintenance, or retraction of the string.

The upper computer generates and wirelessly transmits control commands to both the robotic arm and Lasso Gripper. The system operates through distinct stages: launch, maintain, and retract. During the launch phase, two of the coreless motors are responsible for launching the string inward, while the other two manage the outward launch. The onboard MCU controls both the speed and direction of the string's movement, enabling precise control over the gripper's actions. If the outward rotation speed surpasses the inward rotation speed, more string is released than retracted, enabling the gripper to expand its capture range. During rapid retraction, the string will accumulate and entangle within the gripper. To prevent this, an additional motor is positioned at the center of the Lasso Gripper to wind the string as it is retrieved. This motor serves two primary functions: (1) Organizing the retracted string during the retraction phase. (2) Securing the string in place when the other four motors are switched off. During the release process, all five motors are switched off, allowing the object to fall freely under gravity. 

A piece of cotton string is used as the lasso for our experiment. The reason for the choice is that the cotton string is lightweight to generate enough drag under high speed and its ability to maintain a reasonable surface friction on both the launching wheels and the target.

\section{DYNAMICS AND KINEMATICS ANALYSIS}

With the popularity of string shooter toys on the Internet, some fluid mechanics scholars have been attracted to study them. On the basis of their research, the physical expression of the string loop curve is substituted into the working space of the robotic arm for further analysis.

\subsection{Dynamics of the string loop}

The dynamics of the string loop in Lasso Gripper are rooted in principles of fluid mechanics and physics \cite{abello2024string, daerr2019charmed, taberlet2019propelled}. This section explores the behavior of the string loop as it transitions from a low-velocity, gravity-dominated state to a high-velocity, self-supporting configuration, which facilitates fast grasping using Lasso Gripper.

At low velocities, the string loop is primarily influenced by gravity, causing it to hang downwards. This gravity-dominated regime is characterized by a noticeable sag in the string due to its weight. As the velocity increases, the string undergoes a transition to a high-velocity regime where it forms a self-supporting loop. In this regime, the string is not suspended due to inertia but rather through the aerodynamic drag. This drag force exerted by the surrounding air counteracts gravity, allowing the string to maintain a stable loop shape.

\begin{figure}[htbp]
    \centering
    \includegraphics[width=0.5\linewidth]{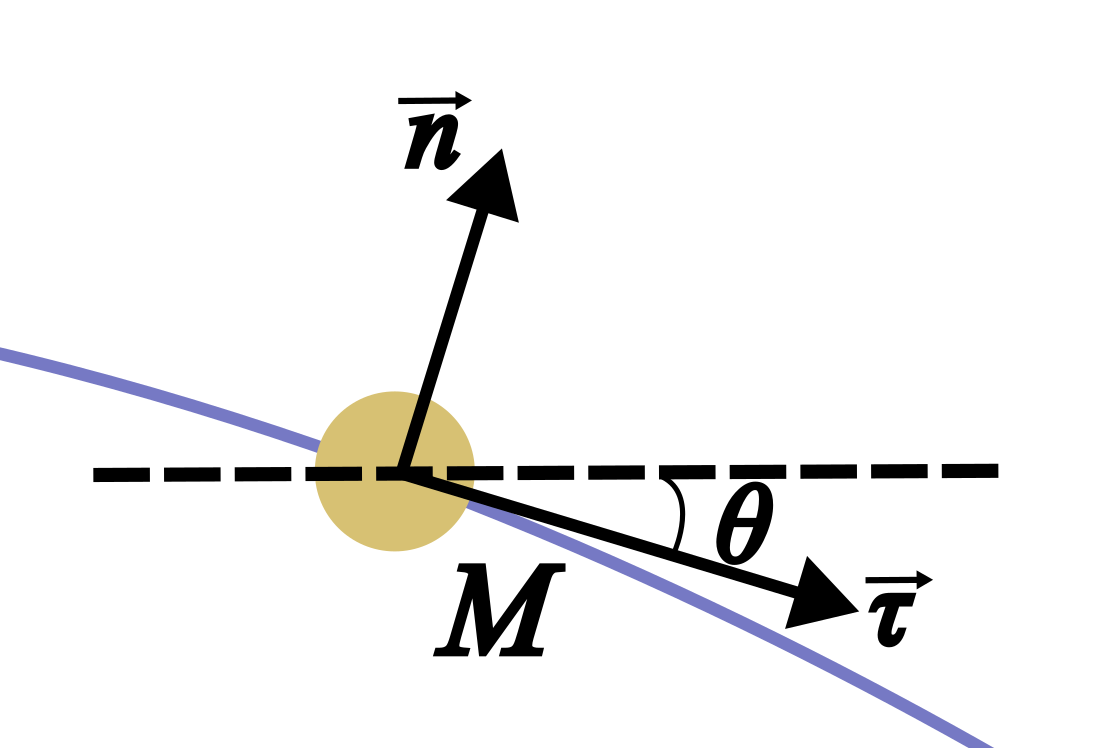} 
    \caption{The 2D local Frenet-Serret frame is defined on minimal model point M \cite{abello2024string}. The unit vector $\vec {\tau}$ represents the curve's tangent and indicates the direction of movement. $\vec {n}$  is the normal unit vector. The angle between the vector $\vec {\tau}$ and the horizontal line is noted as $\theta$.}
    \label{fs}
\end{figure}

Assuming the loop junction is at position $s = s_0$ when $t = 0$, the position of the junction along this curve is $s = v \cdot t + s_0 (\mathrm{mod} \, L) $. $T(s)$ is noted as the tension. As depicted in Fig. \ref{fs}, Frenet-Serret frame $(\vec {\tau}, \vec {n})$ is used to describe the dynamics of an infinitesimal piece of string of length $\mathrm{d}s$ subjected to its weight $\mu \mathrm{d} s \vec {g}$, its tension $\frac{\mathrm{d}(T \vec {\tau})}{\mathrm{d}s}\mathrm{d}s^3$ and the linear air drag $-f \vec {\tau} \mathrm{d} s$ where $f>0$.

Newton's motion equation produces:
\begin{equation}
\label{Newton}
\begin{split}
\frac{\mathrm{d}(\mu v \vec {\tau})}{\mathrm{d}t} = \mu \vec {g} + \frac{\mathrm{d}(T \vec {\tau})}{\mathrm{d}s} - f \vec {\tau}
\end{split}
\end{equation}
Noting that $\mathrm{d}s = v \cdot \mathrm{d}t$, defining the effective tension $T^o = T-\mu v^2$, and recalling that $\vec {\tau} = \frac{\mathrm{d}\vec {\tau}}{\mathrm{d}s}$, we get:
\begin{equation}
\begin{split}
T^o \vec {\tau} - f \vec {\tau} +\mu \vec {g}s=\vec {0}
\end{split}
\end{equation}
If we  appropriately chose the origin for $\vec {r}$ as the point $o$  where the tangent to the loop is vertical, at the end point of the lasso loop, the integration constants will vanish. Defining $tan \theta = \frac{\mathrm{d}z}{\mathrm{d}x}$, we can obtain:
\begin{equation}
\label{Final}
\begin{split}
-f \cdot ( x \mathrm{tan} \theta - z )+\mu gs=0
\end{split}
\end{equation}

\begin{figure}[htbp]
    \centering
    \includegraphics[width=0.7\linewidth]{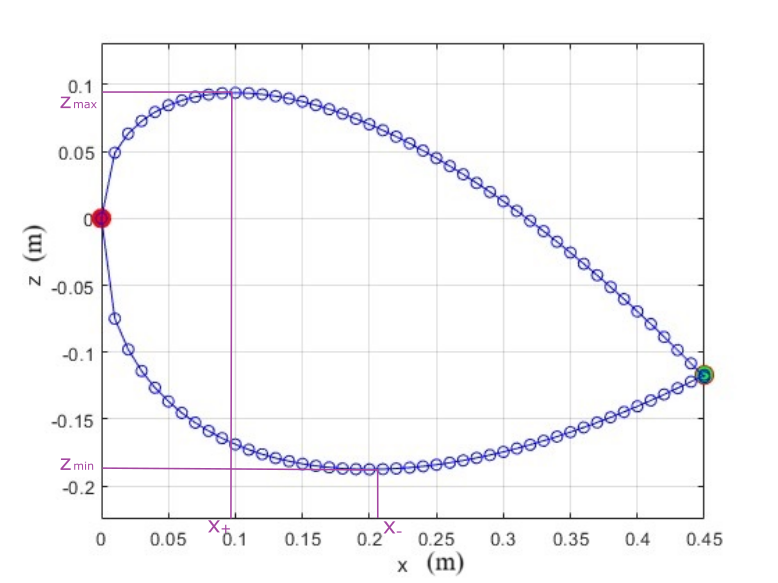} 
    \caption{The numerical solution of loop curve. To facilitate the solution of Eq. \ref{Final}, the origin is set to the position of the red dot where $\frac{\mathrm{d}z}{\mathrm{d}x} = 0$. The physical ejection point is marked as a green point. The top half curve between red point and green point is noted as $z_{+}(x)$, and the bottom half of the curve is noted as $z_{-}(x)$.}
    \label{curve}
\end{figure}

This integro-differential equation can be solved analytically. We get the relationship between z and x in a vertical 2d space in Eq. \ref{Solution}. $z_{+}(x)$ and $z_{-}(x)$ are the top and bottom halves of the loop, respectively. $x_{+}$ and $x_{-}$ are geometry parameters represent for the horizontal distance from $z_{max}$ and $z_{min}$ to original point in Fig. \ref{curve}. $\frac{\mu g}{f}$ is noted as $R$ for convenience.
\begin{equation}
\label{Solution}
\begin{split}
z_{\pm}(x) = \frac{1}{2} \left( 
\frac{x}{1 \pm R}{\left|\frac{x}{x_{\pm}}\right|}^{\pm R} -
\frac{x}{1 \mp R}{\left|\frac{x}{x_{\pm}}\right|}^{\mp R}
\right)
\end{split}
\end{equation}
It is imperative to assert that $R<1$ to ensure the existence of a finite mathematical solution, indicating that the string loop can maintain its configuration only if aerodynamic drag surpasses gravitational forces. The establishment of the self-supporting loop is dictated by the aerodynamic drag exerted on the string. This drag force emerges from the interaction between the moving string and the surrounding air, generating a pressure differential that lifts and supports the string loop.
\begin{figure}[htbp]
    \centering
    \includegraphics[width=0.6\linewidth]{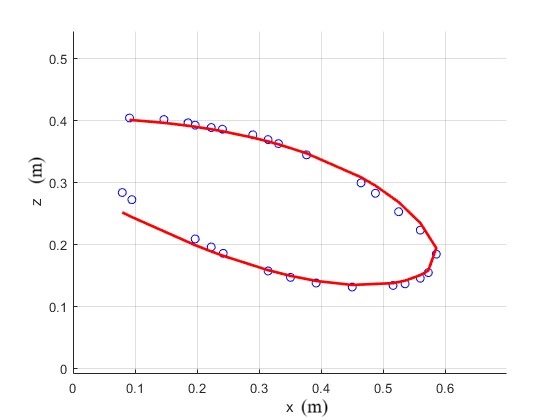} 
    \caption{The theoretical curve is verified against the sampling points.}
    \label{iou}
\end{figure}
In Fig. \ref{iou}, the blue dots represent for the sampling points on the actual string loop measured in Fig. \ref{speed} (B). Fig. \ref{speed} (B) is chosen as it represents a typical middle-speed scenario, which sufficiently validates the theoretical model.  The red curve is drawn according to Eq. \ref{Solution}, where the parameter $[R, x_{+}, x_{-}]$ is adjusted as $[0.33, 0.61m, 0.12m]$. According to the X-coordinate of the experimental sampling points, a corresponding set of points is selected on the theoretical curve. The two sets of points form an effective grasping range, respectively. The intersection of the union between these two ranges is calculated to be 92.23\%. The error between the experimental value and the theoretical curve mainly comes from the direction constraint of the retraction wheels to the string, which is different from the modeling assumption.

In particular, when part of the loop is in contact with a table or wall, the supporting force of the table will change the shape of the loop without affecting the retention of the top half of the loop. This characteristic allows Lasso Gripper to maintain space inside the circle in a restricted environment and maintain satisfactory working conditions.

\subsection{The workspace analysis}
Due to the self-supporting loop configuration of the Lasso Gripper in three-dimensional space, the operational workspace of the proposed manipulator significantly exceeds the physical reach of the robotic arm.

\begin{figure}[ht]
    \centering
    \includegraphics[width=0.8\linewidth]{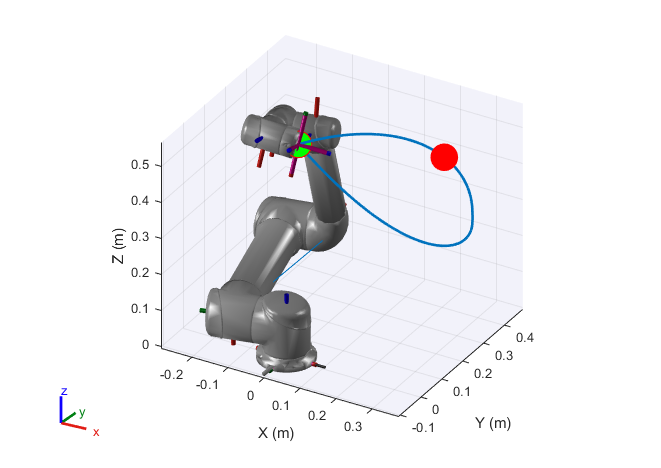} 
    \caption{Lasso Gripper string loop is set at the end of UR5 robotic arm. The green point at the end of the robotic arm marks the position reachable by a rigid manipulator, while the red point indicates the furthest position on the string loop from the original point at the same robotic position.}
    \label{ws1}
\end{figure}

As delineated in Section III-A, the three-dimensional pose kinematics of the lasso-type end-effector can be characterized by constrained dynamics under gravitational loading. For the local coordinate system constrained by gravity, we define the working plane as the plane formed by the lasso's ejection direction and the gravity direction. The rotation with the normal vector of this working plane as the rotation axis is defined as pitch, while the rotation with the gravity vector as the rotation axis is defined as yaw. According to the right-hand coordinate system, the roll rotation axis is defined as the axis orthogonal to both the pitch and yaw rotation axes, lying within the working plane. Specifically, the gripper mechanism exhibits negligible roll angular displacement within the global coordinate frame due to gravitational stabilization. Its pitch orientation becomes parametrically dependent on the projectile's ejection velocity vector, while yaw angle variations are governed by the robotic manipulator's pose in Cartesian space.

\begin{figure}[htp]
    \centering
    \includegraphics[width=0.6 \linewidth]{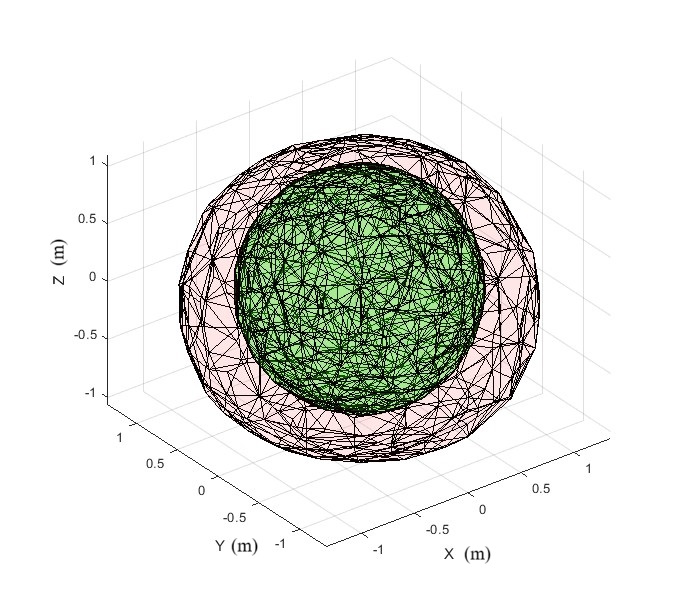} 
    \caption{The initial robotic arm workspace is marked in green and extended Lasso Gripper workspace is marked in red.}
    \label{ws2}
\end{figure}

We conduct a case study with specific parameters to investigate the working space of the Lasso Gripper on a 6-axis robotic arm using the Monte Carlo method. A UR5 robot without joint angle limitation is applied as an example in Fig. \ref{ws1}, and $[R, x_{+}, x_{-}]$ is set as $[0.7, 0.2m, 0.1m]$ in this case. The green point at the end of the robotic arm is the specific position that a rigid manipulator can reach, and the red point represents for the furthest point to the original point on the string loop at the same robotic position. The set of green points and red points in Fig. \ref{ws2} is the workspace of the robotic arm and the extended lasso gripper, respectively. Generating 200,000 sampling points for robotic arm configuration, the volume of workspace convex hull is calculated. The volume of the original UR5 workspace convex hull is $3.4620 m^3$ , and the volume of the extended Lasso Gripper workspace convex hull is $8.9002 m^3$. The workspace extending ratio is up to $157.08\%$.

\section{LASSO GRIPPER GRASPING}

A wide variety of tests in several demonstrations is conducted for Lasso Gripper. The proposed gripper shows the adaptability of the complex grasping environment. Some comparisons between Lasso Gripper and the antipodal point gripper were also studied.

\subsection{Demonstrations}
\begin{figure}[htbp]
    \centering
    \includegraphics[width=0.7 \linewidth]{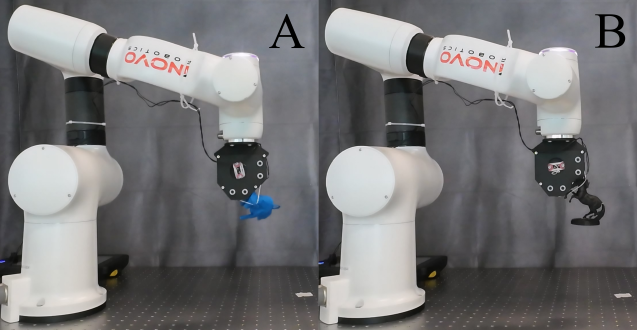} 
    \caption{Validation of grasping ability via bull and horse figures.}
    \label{fig:3}
\end{figure}

\begin{figure}[htbp]
    \centering
    \includegraphics[width= 0.7 \linewidth]{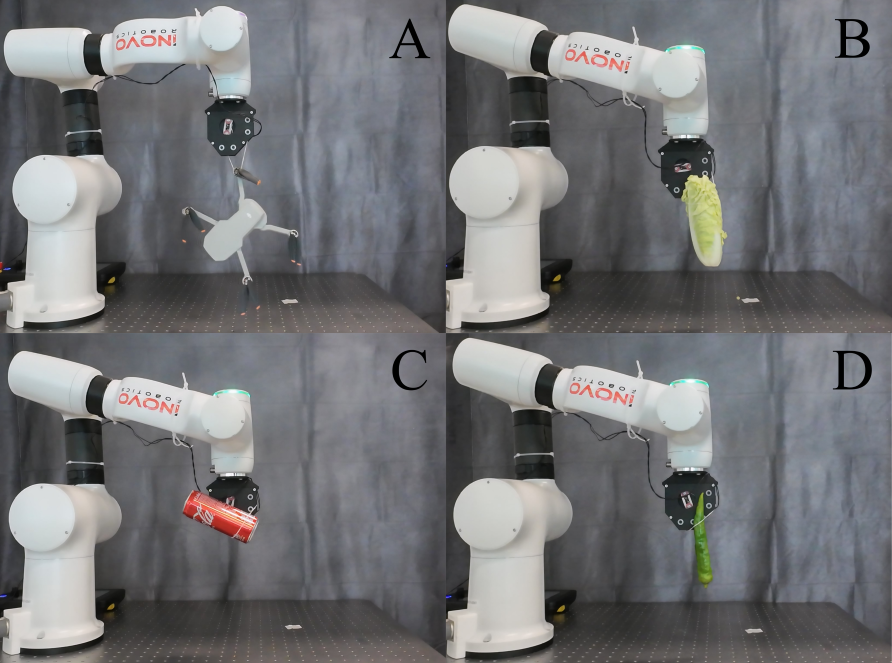} 
    \caption{Validation of grasping ability over variable shapes: drone, cabbage, coke can and chili pepper.}
    \label{fig:4}
\end{figure}

\begin{figure}[htbp]
    \centering
    \includegraphics[width=0.7 \linewidth]{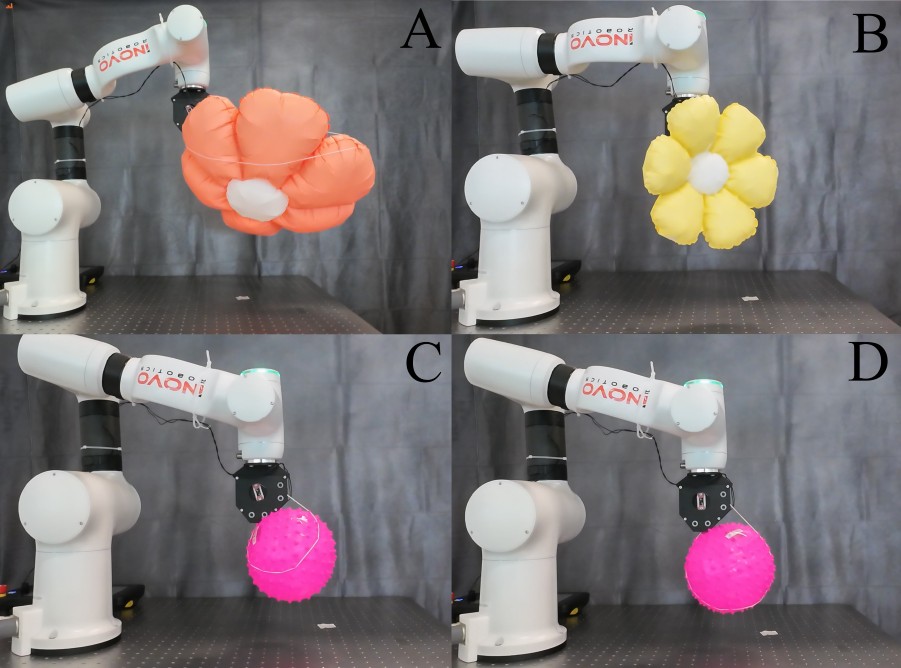} 
    \caption{Validation of grasping ability with oversized objects: inflated balloons.}
    \label{fig:5}
\end{figure}

To demonstrate the unique ability of Lasso Gripper and make comparisons against antipodal ones, the following experiment scenario is set. The Robotic arm used in the experiment is a modular robotic arm from Inovo Robotic with a maximum payload as 6kg. The robotic arm is mounted to a flat platform where the target objects are also placed at. 

The goal of the experiment is not only to validate that the Lasso Gripper performs the core functions of a lasso and an uurga but also to demonstrate its advantages over antipodal grippers in specific tasks. The tests are categorized as follows: 1) Capturing animal figures by the neck and horns; 2) Capturing objects of various shapes; 3) Capturing oversized objects by adapting to their shape; 4)  Capturing moving objects.

For Goals 1 to 3, the target objects were placed on the same table as the robotic arm. The experiment proceeded as follows:
1) The robotic arm moved to the designated position, initiating the launch phase; 2) It then swept the Lasso Gripper toward the target while the gripper maintained its form; 3) Upon reaching the target, the gripper retracted the string to secure the object; 4) The arm lifted vertically while ensuring a firm grip.

 For Goal 4, the robotic arm remained stationary while the Lasso Gripper performed all actions, trying to capture the flying object.

Fig. \ref{fig:3} illustrates the experiments result for Aim 1. The lasso was successfully attached to the horn and the neck, respectively, and firmly grasped them towards the tip of the robotic arm, transporting the figures from the platform. For the second set of target objects, a drone, a cabbage, an empty coke can and a chili pepper were selected. In Fig. \ref{fig:4}, Lasso Gripper successfully grasped the targets and carried it away from the base. 
\begin{figure}[htbp]
    \centering
    \includegraphics[width=0.7 \linewidth]{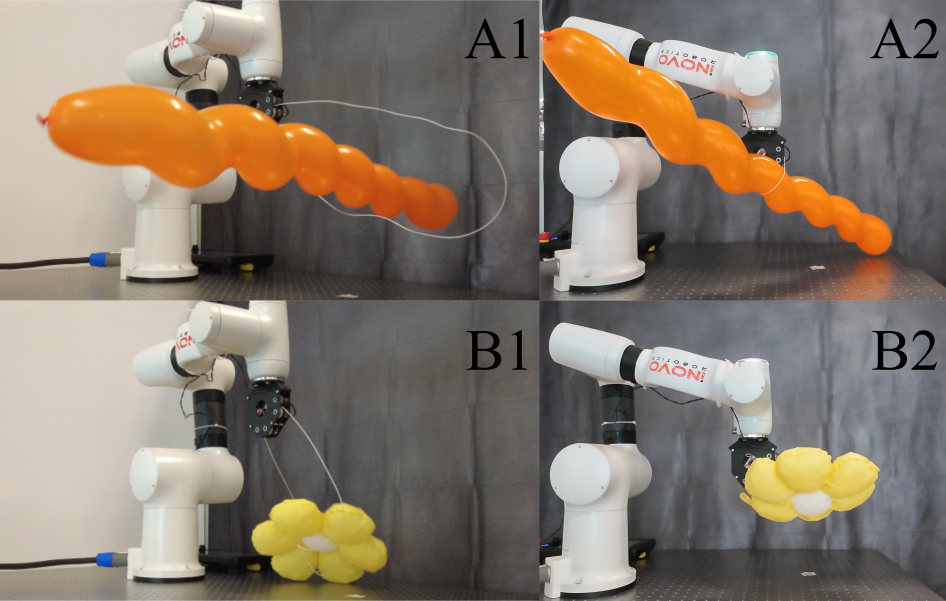} 
    \caption{Flying object is captured by Lasso Gripper.}
    \label{fig:6}
\end{figure}
The grasping strategy for these two experiments is similar. For both cases, the robotic swept right above the target while Lasso Gripper maintained the string in full range, ensuring that the target could be captured as the string retracted. This grasping strategy can also be extended to other expansive grasping targets, such as oversized balloons. Hence, the third set of objects are two flower-shaped balloons, with diameters of 45cm and 71cm, respectively, and a spike ball with a diameter of 18cm, as shown in  Fig. \ref{fig:5}. Though Lasso Gripper could not fully grasp the balloons due to the limitation in the string, it still firmly secured the balloons and lifted off the platform. This laid the foundation for the fourth case of the experiment, which was to capture flying objects. Fig. \ref{fig:6} presented the experiment by manually throwing one long balloon and the small flower balloon through the string made by Lasso Gripper.

Some interesting phenomena were observed during the demonstration, revealing more advantages of Lasso Gripper. Firstly, the string's sustained high speed makes it possible to grasp objects close to the table. In Fig. \ref{fig:4} (B) and Fig. \ref{fig:4} (D), where the cabbage and chili pepper are put very close to the table, Lasso Gripper can still maintain the loop shape and achieve the grasping task. The same goes for the demonstration near a wall. Secondly, the adaptive force closure of the string allows a certain degree of deviated grasping position to be tolerated. Compared with the grasping attitude in Fig. \ref{fig:5} (D), the string loop pose in Fig. \ref{fig:5} (C) occurs an unexpected deviation. Lasso Gripper realizes an adaptive force-closure grasping.

\subsection{Comparisons with Antipodal Point Gripper}
\begin{figure}[htbp]
    \centering
    \includegraphics[width=0.7 \linewidth]{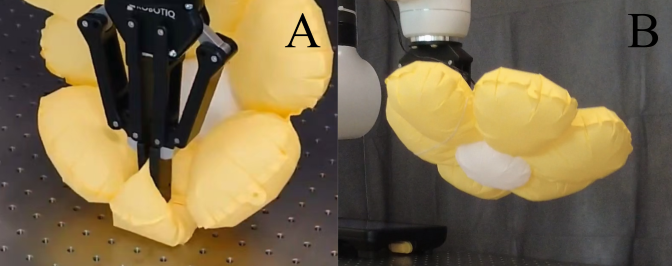} 
    \caption{The inflated balloon is captured by the antipodal gripper. As a result of the concentrated stress applied, the balloon was severely deformed.}
    \label{fig:7}
\end{figure}
As the comparison group, the 2F-140 gripper from Robotiq was used as the classical antipodal gripper. In this scenario, the small flower balloon was used as the target due to its elasticity under external forces. As demonstrated in Fig. \ref{fig:7}, the flower balloon was clearly under dramatic deformation as the antipodal gripper tried to pick it up from the platform. Though the balloon remained intact after the experiment instead of an explosion, the limitations of traditional grippers on handling delicate or plastic objects are straightforward. Instead of directly applying forces and pressure over the surface of the target, Lasso Gripper captured objects via rope tension, avoiding causing damage to the objects. Moreover, this unique feature enables the Lasso Gripper to capture larger and potentially heavier objects than traditional grippers.

\section{CONCLUSIONS AND FUTURE WORK}

This study has introduced a novel lasso gripper design, emphasizing its unique hardware structure and potential applications. The use of a dynamic string loop mechanism offers a flexible and adaptive solution for gripping objects of various shapes and sizes, making it a significant advancement in the field of object manipulation. String loop dynamics were analyzed to determine the shape of the curve. Lasso Gripper's workspace has been proven to have advantages over traditional rigid manipulators. The experiments validate that Lasso Gripper has good adaptability and a satisfactory grasping effect for objects with different shapes and materials.

In future work, we will further optimize the hardware design, explore the influence of factors such as changing the angle of ejection and retraction, changing the material and length of the string. Future work also includes designing a set of unique sensing and grasping trajectory planning algorithms based on the physical characteristics of the rope loop. In addition, by extending the dimension of the string loop and placing multiple Lasso grippers side-by-side at the end of the robotic arm, it is possible to achieve a more stable and fine-controlled grasp.

\bibliographystyle{unsrt}
\bibliography{lasso_gripper}

\end{document}